\documentclass{article}
\usepackage{ijcai21}

\usepackage{times}
\usepackage{soul}
\usepackage{url}
\usepackage[hidelinks]{hyperref}
\usepackage[utf8]{inputenc}
\usepackage[small]{caption}
\usepackage{graphicx}
\usepackage{amsmath}
\usepackage{amsthm}

\usepackage{amssymb}
\usepackage{bm}
\usepackage{booktabs}
\usepackage{algorithm}
\usepackage{algorithmic}
\usepackage{subfigure}
\usepackage{multirow}
\usepackage{latexsym}

\title{CogTree: Cognition Tree Loss for Unbiased Scene Graph Generation}

\author{
Jing Yu$^{1}$\footnote{Equal contribution.}\footnote{Corresponding author.}\and
Yuan Chai$^{2*}$\and
Yujing Wang$^{3}$\and
Yue Hu$^{1}$\And
Qi Wu $^{4}$\\
\affiliations
$^1$Institute of Information Engineering, Chinese Academy of Sciences, Beijing, China\\
$^2$Intelligent Computing and Machine Learning Lab, School of ASEE, Beihang University, Beijing, China \\
$^3$Key Laboratory of Machine Perception, MOE, School of EECS, Peking University, Beijing, China\\
$^4$University of Adelaide, Australia\\
\emails
\{yujing02, huyue\}@iie.ac.cn,
chaiyuan@buaa.edu.cn,
yujwang@pku.edu.cn,
qi.wu01@adelaide.edu.au
}

\begin{document}

\maketitle
\begin{abstract}

Scene graphs are semantic abstraction of images that encourage visual understanding and reasoning. However, the performance of Scene Graph Generation (SGG) is unsatisfactory when faced with biased data in real-world scenarios. Conventional debiasing research mainly studies from the view of  balancing data distribution or learning unbiased models and representations, ignoring the correlations among the biased classes. 
In this work, we analyze this problem from a novel cognition perspective: automatically building a hierarchical cognitive structure from the biased predictions and navigating that hierarchy to locate the relationships, making the tail relationships receive more attention in a coarse-to-fine mode. To this end, we propose a novel debiasing Cognition Tree (CogTree) loss for unbiased SGG. We first build a cognitive structure CogTree to organize the relationships based on the prediction of a biased SGG model. The CogTree distinguishes remarkably different relationships at first and then focuses on a small  portion  of  easily  confused  ones. Then, we propose a 
debiasing loss specially for this cognitive structure, which supports coarse-to-fine distinction for the correct relationships. 
The loss is model-agnostic and  consistently boosting the performance of several state-of-the-art models.  The code is available at: https://github.com/CYVincent/Scene-Graph-Transformer-CogTree.
\end{abstract}

\begin{figure}[t]
    \centering
    \subfigure[Flat thinking.]{
        \includegraphics[width=0.9\columnwidth]{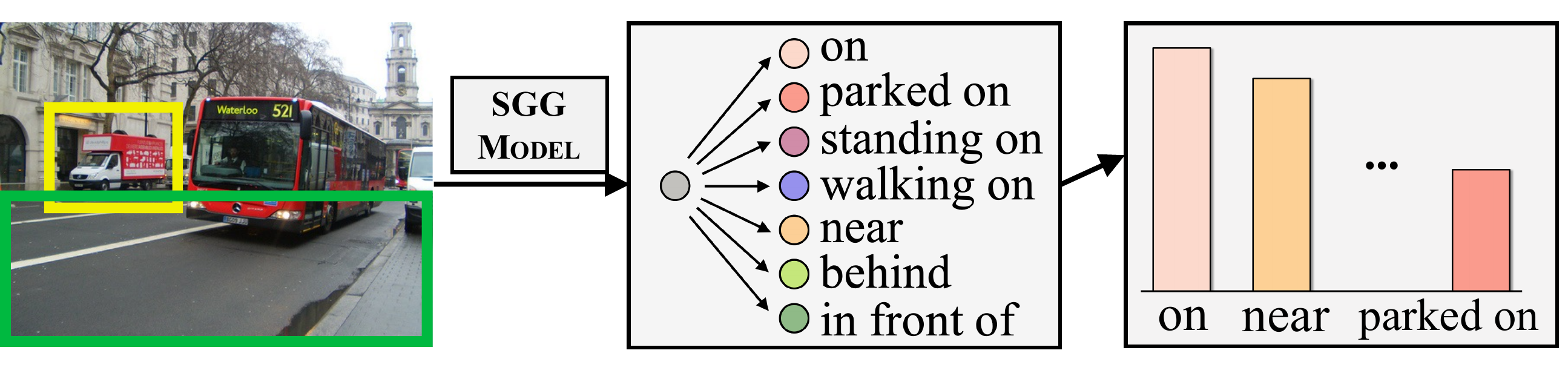}
    }
    \subfigure[Cognition-based hierarchical thinking.]{
        \includegraphics[width=0.9\columnwidth]{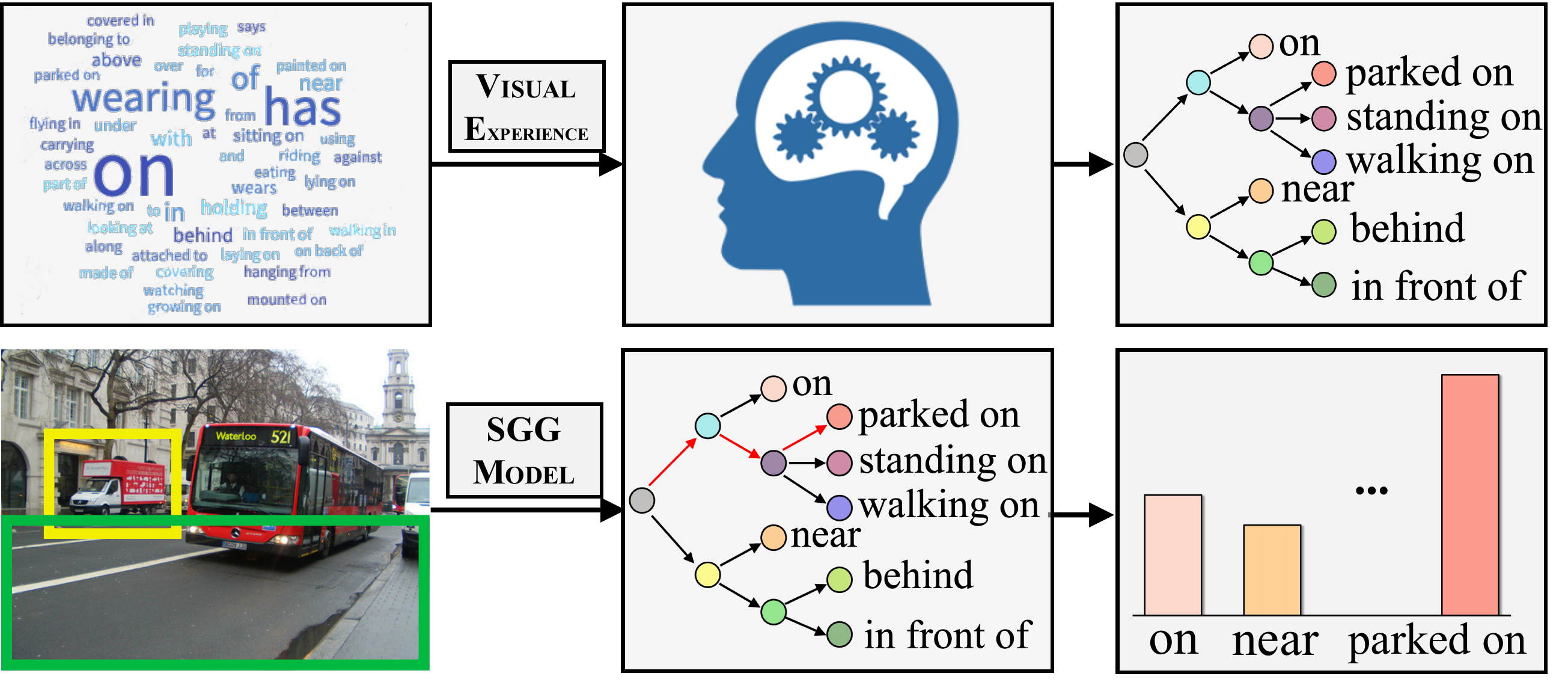}
    }
    \caption{(a) SGG model with conventional flat loss. (b) SGG model with our proposed cognition tree loss. The word size in the top-left box is proportional to the relationship frequency.}
    \label{fig:comparison}
\end{figure}

\section{Introduction}
\label{sec:intro}
Making abstraction from an image into high-level semantics is one of the most remarkable capabilities of humans. 
Scene Graph Generation (SGG) \cite{xu2017scene} $-$ a task of extracting objects and their  relationships in an image to form a graphical representation $-$ aims to achieve the abstraction capability and bridge the gap between vision and language. SGG has greatly benefited the down-stream tasks of question answering \cite{norcliffe2018learning,Zhu2020Mucko} and visual understanding \cite{shi2019explainable,Jiang2020DualVD}. Some works \cite{Zhu2020Mucko,Jiang2020DualVD} feed the scene graphs into graph neural networks for relation-aware object representation. Some others \cite{hudson2019learning} perform sequential reasoning by traversing the relational graphs. Compared with independent objects, the rich relationships 
benefit  explainable reasoning.


Although much effort has been made  with high accuracy in object detection, the detected relationships are far from satisfaction due to the long-tailed data distribution. Only a small portion of the relationships have abundant samples (head) while most ones contain just a few (tail). This heavily biased training data causes biased relationship prediction. Tail relationships will be mostly predicted into head classes, which are not  discriminative  enough for high-level reasoning, such as falsely predicting {\ttfamily on} instead of {\ttfamily looking at}, and coarsely predicting {\ttfamily on} instead of {\ttfamily walking on}.  Besides, tail relationships like {\ttfamily standing on}, {\ttfamily sitting on} and {\ttfamily lying on} can be even harder to distinguish from each other due to their visual similarity and scarce training data.

To tackle this problem, most research focuses on 
learning unbiased models by re-weighting losses \cite{zareian2020bridging} or disentangling unbiased representations \cite{Tang2020Unbiased}. However, humans can effectively infer the correct relationships even when some relationships appear more frequently than others. The essential difference between human and AI systems that has been ignored lies neither in the learning strategy nor feature representation, but in the way that the concepts are organized. 
To illustrate this discrepancy, we show an example {\ttfamily truck parked on street} in Figure \ref{fig:comparison}. Existing models treat all the relationships independently for flat classification (Figure \ref{fig:comparison}(a)). Because of the data bias, the head relationships, \textit{e.g.} {\ttfamily on} and {\ttfamily near}, obtain high predicted probability. In contrast, the cognition theory \cite{Sarafyazd2019Hierarchical} supports that humans process information hierarchically in the  prefrontal  cortex  (PFC). 
We intuitively start from making a rough distinction between remarkably different relationships. As shown in Figure \ref{fig:comparison}(b), relationships belonging to the concept  ``on'',
\textit{e.g.} {\ttfamily on} and {\ttfamily parked on}, will firstly be distinguished from the ones about ``near'', \textit{e.g.} {\ttfamily near} and {\ttfamily behind}; then we move to distinguish the slight discrepancy among easily confused ones in one concept, 
\textit{e.g.} {\ttfamily parked on}, {\ttfamily standing on} and {\ttfamily walking on}.

Inspired by the hierarchical reasoning mechanism in PFC, we propose a  novel loss function, \textbf{Cognition Tree} (\textbf{CogTree}) loss, for unbiased scene graph generation. We first propose to build a hierarchy of the relationships, imitating the knowledge structure built from the independent relationships in human mind. The CogTree is derived from the prediction of a biased SGG model that satisfies the aforementioned  thinking principles: distinguishing remarkably different relationships at first and then 
focusing on a small portion of easily confused ones. Then we design a CogTree-based loss to train the SGG network from scratch. This loss enables the network to surpass the noises from inter-concept relationships and then intra-concept relationships progressively. It frees the SGG models from the burden of distinguishing detailed discrepancy among all the relationships at one time, resulting in 
more accurate prediction via this coarse-to-fine strategy.  


The main contributions are summarized as follows: (1) We propose a debiasing loss based on the inductive hierarchical structure,  which supports coarse-to-fine distinction for the correct relationships while progressively eliminating the interference of irrelevant ones. (2) We automatically build the cognitive structure of the relationships from the biased SGG predictions, which reveals the inherent hierarchy that the relationships are organized after biased training; 
 (3) The proposed loss is model-agnostic and can be applied to several representative SGG models. The results of extensive experiments indicate that the CogTree loss remarkably alleviates the bias in SGG and our loss achieves a new state-of-the-art performance compared with existing debiasing methods. 

\section{Related Work}
\label{sec:relatedWork}
\begin{figure*}[ht]
    \centering
    \includegraphics[width=\textwidth,]{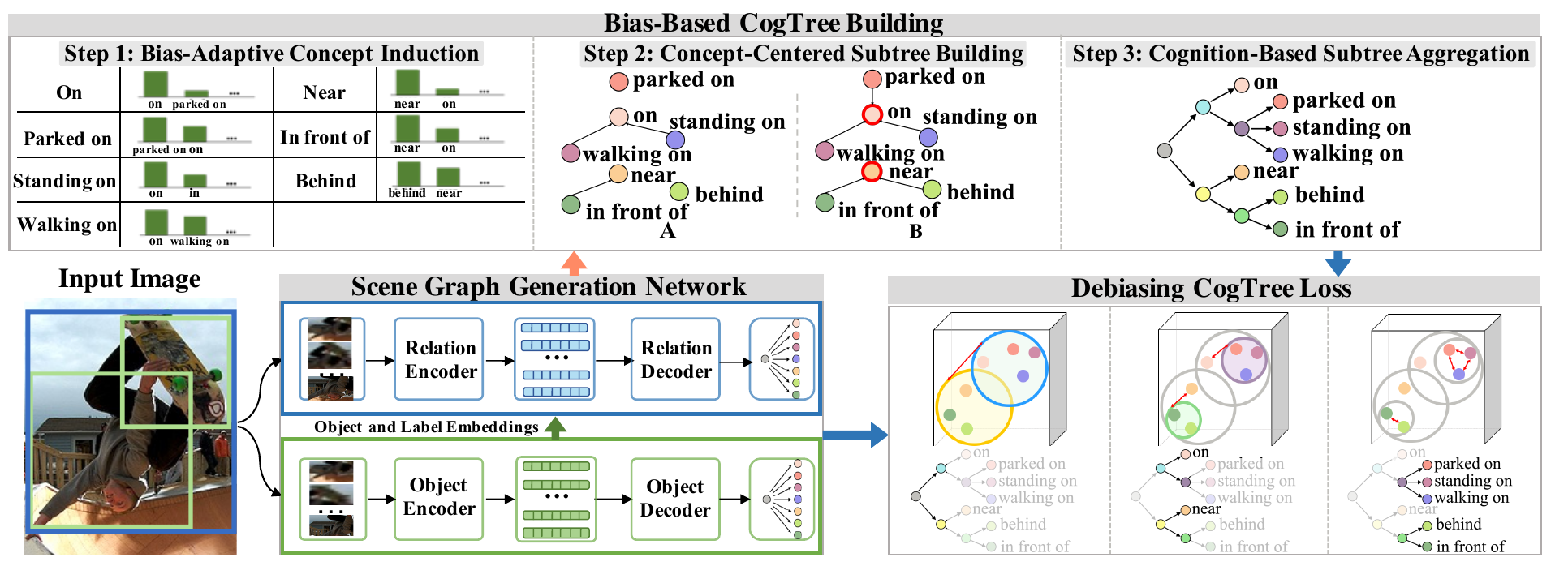}
    \caption{The overview of CogTree loss applied to SGG models. It contains three parts: Scene Graph Generation Network summarizes the framework of biased SGG models; Bias-Based CogTree Building organizes relationships by a coarse-to-fine tree structure based on biased predictions; Debiasing CogTree Loss supports network to distinguish relationships hierarchically. }
    \label{fig:framwork}
\end{figure*}

\textbf{Scene Graph Generation.}
SGG \cite{xu2017scene} products graphical abstraction of an image and encourages visual relational reasoning and understanding in various down-stream tasks \cite{shi2019explainable,Zhu2020Mucko}. Early works focus on object detection and relationship detection via independent networks \cite{lu2016visual,Zhang2017Relationship}, ignoring the rich contextual information. To incorporate the global visual context, recent works leverage message passing mechanism \cite{xu2017scene,chen2019counterfactual},  
recurrent sequential architectures \cite{zellers2018neural,tang2019learning} and contrastive learning \cite{Zhang2019Graphical} 
for more discriminative object and relationship representations. Although the accuracy is high in object detection, the relationship detection is far from satisfaction due to the heavily biased data. \cite{chen2019knowledge,tang2019learning} consider the biased SGG problem and propose mean Recall as the unbiased metric without corresponding debiasing solutions. The recent work \cite{Liang2019VRR} prunes the predominant relationships and keeps the tail but informative ones in the dataset. \cite{Tang2020Unbiased} proposes the first solution for unbiased SGG  by counterfactual surgeries on causal graphs. We rethink SGG task from the cognition 
view and novelly  solve the debiasing problem based on the coarse-to-fine structure of the relationships. 
\\
\textbf{Biased Classification.}
Classification on highly-biased training data has been extensively studied in previous work, which can be divided into three categories: (1) balancing data distribution by data augmentation or re-sampling 
\cite{burnaev2015influence,li2019repair}; 
(2) debiasing learning strategies by re-weighting losses or training curriculums \cite{lin2017focal,Cui2019Class}; 
(3) separating biased representations from the unbiased for prediction \cite{cadene2019rubi,Tang2020Unbiased}. 
Our CogTree loss belongs to the second category but differs from existing methods in that we are the first to leverage the hierarchical structure inherent in the relationships for re-weighting, which enables more discriminative representation learning by a coarse-to-fine mode.

\section{Methodology}
\label{sec:method}

We first introduce the automatic CogTree building process. 
Then a novel debiasing loss based on the tree structure is proposed to train the SGG models. 
Since the CogTree loss is model-agnostic and applicable to various SGG models, 
we case-study two representative models, \textit{i.e.} the widely compared MOTIFS \cite{zellers2018neural} and the state-of-the-art VCTree \cite{tang2019learning}, and a newly proposed model, which first applies the transformer \cite{vaswani2017attention} architecture to the SGG scenario and achieves superior performance. We name it as SG-Transformer and take it as a strong baseline.
The framework is illustrated in Figure \ref{fig:framwork}.

\subsection{Automatic CogTree Building}
\label{ssec: buildTree}

Once the SGG model has been trained on the long-tailed data, the model tends to be biased on predicting ``head'' relationships. 
When different samples are predicted with the same relationship, they mostly share similar properties on either visual appearance (\textit{e.g.} {\ttfamily walking on} and {\ttfamily standing on}) or high-level semantics (\textit{e.g.} {\ttfamily has} and {\ttfamily with}).
We regard these relationships sharing common properties as in one \textit{concept}.  
We automatically induce concepts from the biased predictions and re-organize the relationships based on the concepts by a tree structure, denoted as Cognition Tree (CogTree). 
As shown on the top of Figure \ref{fig:framwork}, this automatic process contains the following three steps: 
\\
\textbf{Step 1: Bias-Based Concept Induction.} 
For all the samples 
in the ground-truth class $d_i$, 
we predict their labels via a biased model and calculate the distribution of predicted label frequency, denoted as $P_i$. 
The most frequently predicted class $d_j$ is regarded as the \textit{concept relationship} of the ground-truth class $d_i$. 
As shown in Figure \ref{fig:framwork}(Step 1), {\ttfamily on} is the concept relationship of itself, {\ttfamily standing on} and {\ttfamily walking on}. 
The above operation induces all the relationships into $C$ concepts with corresponding concept relationships  $\{c_i\}^C$.  
\\
\textbf{Step 2: Concept-Centered Subtree Building.} We re-organize all the relationships in each concept by a \textit{Concept-Centered Subtree}. 
For the $i^{th}$ subtree, the root is $c_i$ while the leaves are the fine-grained relationships induced in $c_i$. 
Note that, if a subtree only contains a root $c_i$, \textit{e.g.} {\ttfamily parked on} in Figure \ref{fig:framwork} (Step 2(A)), then $c_i$ is considered not  representative enough to summarize common properties. We further induce it to the most approximate concept relationship, which has  the second highest frequency in $P_i$, \textit{e.g.} induce {\ttfamily parked on} to {\ttfamily on} in Figure \ref{fig:framwork} (Step 2(B)). 
This process outputs $T$ subtrees. \\
\textbf{Step 3: Cognition-Based Subtree Aggregation.} We aggregate the $T$ subtrees into one hierarchical CogTree. 
As illustrated in Figure \ref{fig:treeloss}, CogTree contains four functional layers and each layer induces relationships into coarser groups than the layer that comes after it. Specifically, the \textit{Root Layer} ($y_0$) contains a virtual node. The  \textit{Concept Layer} ($y_1$) distinguishes which concept the input belongs to. It contains $T$ virtual nodes, each representing a subtree induced in Step 2. 
The following \textit{Coarse-fine Layer} ($y_2$) tells whether the input can be described by a coarse-grained or fine-grained relationship in a certain concept.  
To this end, $y_2$ splits each node in $y_1$ into two nodes: one leaf node indicating the concept relationship while the other virtual node representing the cluster of fine-grained relationships in that concept. 
Each virtual node in $y_2$ links to its fine-grained relationships of the corresponding concept in the \textit{Fine-grained Layer} ($y_3$). $y_3$ focuses on distinguishing the slight discrepancy among easily confused relationships, \textit{e.g.} {\ttfamily standing on} and {\ttfamily walking on}.

\subsection{Debiasing CogTree Loss}
\label{sec:TreeLoss}
We propose a CogTree loss for training the SGG model for unbiased predictions. 
This loss encourages the model to classify relationships from coarse to fine according to the CogTree structure (see the bottom right of Figure \ref{fig:framwork}). To better make the loss functions (\ref{eq:TCB}) and (\ref{eq:CB}) understood, several notations are provided in advance as below:
\\
\textbf{Ground-Truth CogTree Path (GCP).} It denotes the ground-truth classification path in CogTree for a training sample. Given the ground-truth label of a sample, we track the path from the root to the leaf with that label in CogTree and denote this path as the ground-truth path $L_{path}=\{l_{0}, l_{1}, ... l_{M}\}$, where the $k^{th}$ value $l_{m}$ is the ground-truth node at layer $y_m$.
\\
\textbf{Predicted CogTree Probability (PCP).} It means the predicted probability of each node in CogTree for a given sample by a biased SGG model. We denote the predicted probability set of all the classes  as $P_{pred}$ and denote the probability of class $k$ as $p_k$. The PCP of node $i$ is defined as below:
\begin{equation}
z_i=\begin{cases}
p_k & \text{if } i \text{ is a leaf and } class(\textit{i})=k \\ 
\frac{1}{\left | C(i) \right |} \sum_{j\in C(i)} z_j & \text{if } i \text{ is not a leaf}
\end{cases}
\label{eq:probability}
\end{equation}
where $C(i)$ means the children of node $i$. \textit{class(i}) denotes the corresponding class of node
$i$.
\\
\textbf{Class-Balanced Weight (CBW).}  It denotes the balance weight of each node in CogTree. Since the success of re-weighting strategy for debiasing, we adopt a well performed weighting factor \cite{Cui2019Class} to compute CBW of each leaf and uniformly define CBW of each node as below:
\begin{equation}
w_i=\begin{cases}
(1-\beta)/(1-\beta^{n_i}) & \text{if } i \text{ is a leaf} \\ 
\frac{1}{\left | C(i) \right |} \sum_{j\in C(i)} w_j& \text{if } i \text{ is not a leaf}
\end{cases}
\label{eq:weight}
\end{equation}
where $\beta$ is a hyper-parameter  in the range of $[0,1)$ and $n_i$ is the sample number of the corresponding class of leaf $i$.
\begin{figure}[t]
    \centering
    \includegraphics[width=0.8\columnwidth]{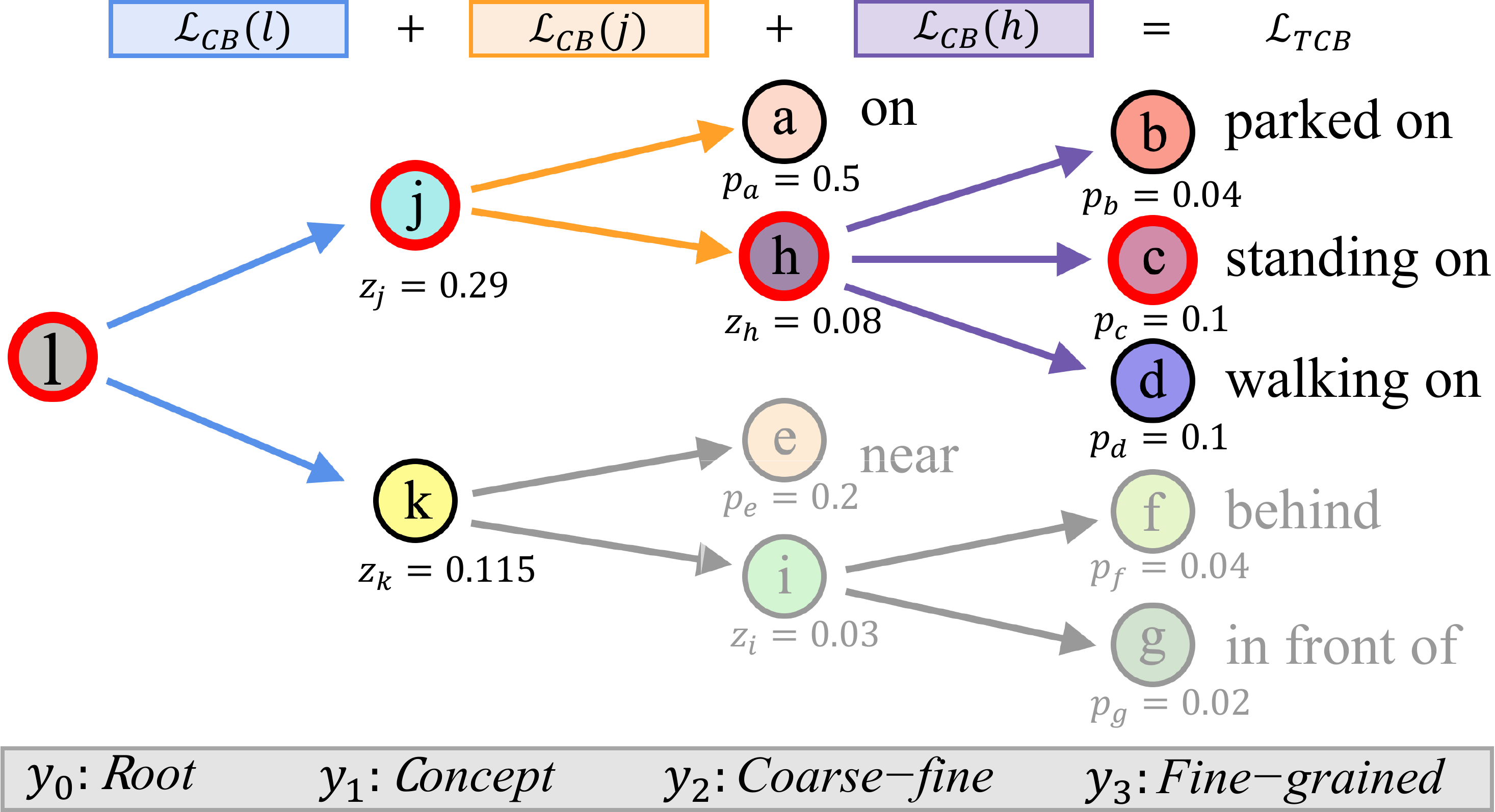}
    \caption{Illustration of calculating the TCB loss.}
    \label{fig:treeloss}
\end{figure}
\\
\textbf{CogTree Loss.} It contains two  parts for complementary benefits: CogTree-based class-balanced (TCB) loss for debiasing by coarse-to-fine relationship classification and  class-balanced (CB) loss for debiasing by re-weighting on softmax cross-entropy loss. The two loss terms are defined below. 

Given a sample with the ground-truth path $L_{path}$, as illustrated in Figure \ref{fig:treeloss}, we compute the class-balanced softmax cross-entropy loss on each layer of CogTree, and average the results to obtain the TCB loss as below:
\begin{equation}
\mathcal{L}_{TCB}=\frac{1}{|L_{path}|}\sum_{i\in L_{path}}-w_{i}\textup{log}(\frac{\textup{exp}(z_{i})}{\sum\nolimits_{z_j\in B(i)}\textup{exp}(z_{j})})
\label{eq:TCB}
\end{equation}
where $B(i)$ means the brothers of node $i$. TCB loss  forces  the  network  to  surpass  the  noises from inter-concept relationships and learn concept-specific embeddings first, and then surpass  the  noises within one concept to refine relationship-specific embeddings,  
resulting in more fine-grained predictions. 

Given a sample with the ground-truth class $k$, CB loss computes the class-balanced softmax cross-entropy based on the biased predicted probabilities $P_{pred}$: 
\begin{equation}
\mathcal{L}_{CB}=-w_{k}\textup{log}(\frac{\textup{exp}(p_{k})}{\sum\nolimits_{p_{j}\in P_{pred}}\textup{exp}(p_{j})})\label{eq:CB}
\end{equation}
where $w_k$ is the weighting factor \cite{Cui2019Class} for class $k$. Empirically, we found it the best to combine TCB with CB. 
In conclusion, our CogTree loss is: 
\begin{equation}
\mathcal{L} = \mathcal{L}_{CB} + \lambda\mathcal{L}_{TCB}
\label{eq:fullLoss}
\end{equation}
where $\lambda=1$ is the balancing weight. The SGG model is trained from scratch by the CogTree loss and predicting results via flat multi-class classifier without modification. 

\subsection{Scene Graph Generation Models}
\label{ssec:biasedModel}

\begin{figure}[t]
    \centering
    \includegraphics[width=0.9\columnwidth]{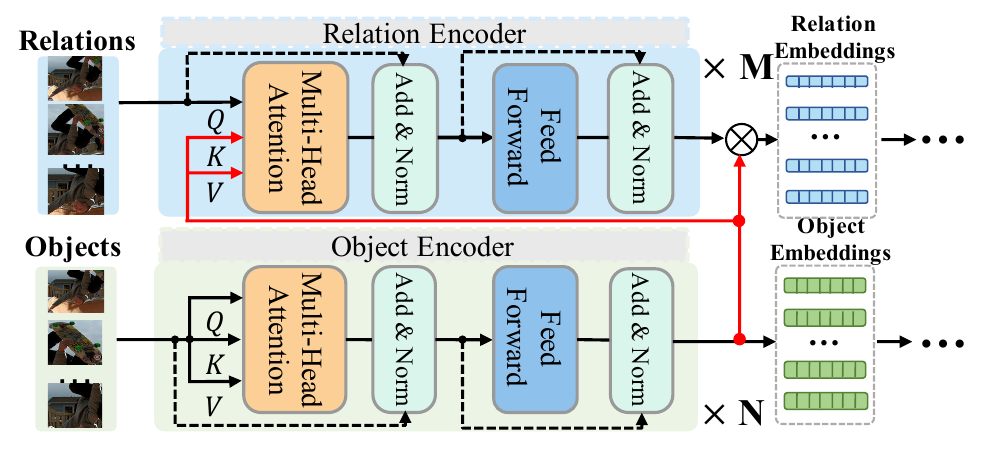}
    \caption{The encoder structure in SG-Transformer.}
    \label{fig:SGTransformer}
\end{figure}

We case-study on 
three SGG models, \textit{i.e.} MOTIFS, VCTree and our SG-Transformer. As shown in the bottom left of Figure \ref{fig:framwork}, their architectures  mainly contains two processes: object classification (bottom)  for detecting the object positions and labels and relationship classification (top) for predicting the relationships between objects. We first review  MOTIFS, and then introduce VCTree and SG-Transformer in a comparable fashion. All models are trained by the conventional cross-entropy loss \cite{zellers2018neural}.
\\
\textbf{MOTIFS.} A pre-trained Faster R-CNN \cite{ren2015faster} is applied to extract top $K$ objects $O=\{o_i\}^K$ and describe object $o_i$ by the RoIAlign visual feature $v_i$,  
the tentative object label $l_i$, and the spatial feature $b_i$. The above three kinds of features are concatenated and fed into $Encoder_o$ to obtain the refined object embedding $m_i$. $m_i$ is then decoded by  $Decoder_o$ to predict its final object label $l_i^o$. MOTIFS implemented $Encoder_o$ as a Bi-LSTM and adopts an LSTM as $Decoder_o$. The relationship encoder $Encoder_r$ first adopts another Bi-LSTM to capture the contextual information and then fuses the pairwise object features with the union region feature for decoding. The relationship decoder $Decoder_r$ is a fully connected layer followed by a softmax layer. 
\\
\textbf{VCTree.} Different from MOTIFS, VCTree utilizes two independent TreeLSTMs as $Encoder_o$ and $Encoder_r$, respectively. It additionally takes the bounding box pair feature as another input of $Decoder_r$.
\\
\textbf{SG-Transformer.} It adopts the  transformer architecture as $Encoder_o$ and  $Encoder_r$ by adjusting inputs for the SGG scenario. As shown in Figure \ref{fig:SGTransformer}, $Encoder_o$ contains $N$ object-to-object (O2O) transformer blocks with the input of object embeddings $\{v_i\}^K$. $Encoder_r$ consists of $M$ relation-to-object (R2O) transformer blocks with the input of relationship emebddings $\{r_{ij}\}^{K\times K}$  and the object embeddings $\{m_i\}^K$. $r_{ij}$ is a combination of  three kinds of features from objects $o_i$ and $o_j$: spatial features, union region feature, word embeddings \cite{pennington2014glove} 
of the predicted object labels. The output from $Encoder_r$ is concatenated with $m_i$ and $m_j$ for decoding. $Decoder_o$ and $Decoder_r$ contain a fully connected layer followed by a softmax layer. 

\begin{table*}[!ht]
\centering
\resizebox{\textwidth}{!}{\scriptsize
\begin{tabular}{lcccccc}
\hline
                                              & \multicolumn{2}{c}{Predicate Classification}                            & \multicolumn{2}{c}{Scene Graph Classification}                         & \multicolumn{2}{c}{Scene Graph Detection}                              \\ \hline
\multicolumn{1}{l|}{Model}                    & mR@20 / 50 / 100            &\multicolumn{1}{c|}{ R@20 / 50 / 100}      & mR@20 / 50 / 100            & \multicolumn{1}{c|}{R@20 / 50 / 100}     & mR@20 / 50 / 100           &R@20 / 50 / 100      \\ \hline
\multicolumn{1}{l|}{IMP+}                      &    - / 9.8 / 10.5           &\multicolumn{1}{c|}{ 52.7 / 59.3 / 61.3 }  &    - /  5.8  / 6.0          & \multicolumn{1}{c|}{31.7 / 34.6 / 35.4}  &   - /  3.8  / 4.8          &14.6 / 20.7 / 24.5   \\
\multicolumn{1}{l|}{FREQ}                     & 8.3  / 13.0 / 16.0          &\multicolumn{1}{c|}{ 53.6 / 60.6 / 62.2 }  & 5.1  /  7.2  / 8.5          & \multicolumn{1}{c|}{29.3 / 32.3 / 32.9}  & 4.5 /  6.1  / 7.1          &20.1 / 26.2 / 30.1   \\
\multicolumn{1}{l|}{KERN}                     &    - / 17.7 / 19.2          &\multicolumn{1}{c|}{    - / 65.8 / 67.6 }  &    - /  9.4  / 10.0         & \multicolumn{1}{c|}{   - / 36.7 / 37.4}  &   - /  6.4  / 7.3          &   - / 27.1 / 29.8   \\
\multicolumn{1}{l|}{MOTIFS}                   & 10.8 / 14.0 / 15.3          &\multicolumn{1}{c|}{ 58.5 / 65.2 / 67.1 }  & 6.3  /  7.7  / 8.2          & \multicolumn{1}{c|}{32.9 / 35.8 / 36.5}  & 4.2 /  5.7  / 6.6          &21.4 / 27.2 / 30.3   \\
\multicolumn{1}{l|}{VCTree}                   & 14.0 / 17.9 / 19.4          &\multicolumn{1}{c|}{ 60.1 / 66.4 / 68.1 }  & 8.2  / 10.1 / 10.8          & \multicolumn{1}{c|}{35.2 / 38.1 / 38.8}  & 5.2 /  6.9  / 8.0          &22.0 / 27.9 / 31.3   \\ \hline
\multicolumn{1}{l|}{MOTIFS*}                  & 11.5 / 14.6 / 15.8          &\multicolumn{1}{c|}{ 59.5 / 66.0 / 67.9 }  & 6.5  /  8.0  / 8.5          & \multicolumn{1}{c|}{35.8 / 39.1 / 39.9}  & 4.1 /  5.5  / 6.8          &25.1 / 32.1 / 36.9   \\
\multicolumn{1}{l|}{MOTIFS + Focal}           & 10.9 / 13.9 / 15.0          &\multicolumn{1}{c|}{ 59.2 / 65.8 / 67.7 }  & 6.3  /  7.7  / 8.3          & \multicolumn{1}{c|}{36.0 / 39.3 / 40.1}  & 3.9 /  5.3  / 6.6          &24.7 / 31.7 / 36.7   \\
\multicolumn{1}{l|}{MOTIFS + Reweight}        & 16.0 / 20.0 / 21.9          &\multicolumn{1}{c|}{ 45.4 / 57.0 / 61.7 }  & 8.4  / 10.1 / 10.9          & \multicolumn{1}{c|}{24.2 / 29.5 / 31.5}  & 6.5 /  8.4  / 9.8          &18.3 / 24.4 / 29.3   \\
\multicolumn{1}{l|}{MOTIFS + Resample}        & 14.7 / 18.5 / 20.0          &\multicolumn{1}{c|}{ 57.6 / 64.6 / 66.7 }  & 9.1  / 11.0 / 11.8          & \multicolumn{1}{c|}{34.5 / 37.9 / 38.8}  & 5.9 /  8.2  / 9.7          &23.2 / 30.5 / 35.4   \\
\multicolumn{1}{l|}{MOTIFS + TDE}             & 18.5 / 25.5 / \textbf{29.1}          &\multicolumn{1}{c|}{ 33.6 / 46.2 / 51.4 }  & 9.8  / 13.1 / 14.9          & \multicolumn{1}{c|}{21.7 / 27.7 / 29.9}  & 5.8 /  8.2  / 9.8          &12.4 / 16.9 / 20.3   \\
\multicolumn{1}{l|}{MOTIFS + CogTree}         & \textbf{20.9 / 26.4} / 29.0 &\multicolumn{1}{c|}{ 31.1 / 35.6 / 36.8 }  & \textbf{12.1 / 14.9 / 16.1} & \multicolumn{1}{c|}{19.4 / 21.6 / 22.2}  & \textbf{7.9 / 10.4 / 11.8} &15.7 / 20.0 / 22.1   \\ \hline
\multicolumn{1}{l|}{VCTree*}                  & 11.7 / 14.9 / 16.1          &\multicolumn{1}{c|}{ 59.8 / 66.2 / 68.1 }  & 6.2  /  7.5  / 7.9          & \multicolumn{1}{c|}{37.0 / 40.5 / 41.4}  & 4.2 /  5.7  / 6.9          &24.7 / 31.5 / 36.2   \\
\multicolumn{1}{l|}{VCTree + TDE}             & 18.4 / 25.4 / 28.7          &\multicolumn{1}{c|}{ 36.2 / 47.2 / 51.6 }  & 8.9  / 12.2 / 14.0          & \multicolumn{1}{c|}{19.9 / 25.4 / 27.9}  & 6.9 /  9.3  / 11.1         &14.0 / 19.4 / 23.2   \\
\multicolumn{1}{l|}{VCTree + CogTree}         & \textbf{22.0 / 27.6 / 29.7} &\multicolumn{1}{c|}{ 39.0 / 44.0 / 45.4 }  & \textbf{15.4 / 18.8 / 19.9} & \multicolumn{1}{c|}{27.8 / 30.9 / 31.7}  & \textbf{7.8 / 10.4 / 12.1} &14.0 / 18.2 / 20.4   \\ \hline
\multicolumn{1}{l|}{SG-Transformer}           & 14.8 / 19.2 / 20.5          &\multicolumn{1}{c|}{ 58.5 / 65.0 / 66.7 }  & 8.9  / 11.6 / 12.6          & \multicolumn{1}{c|}{35.6 / 38.9 / 39.8}  & 5.6 /  7.7  / 9.0          &24.0 / 30.3 / 33.3   \\
\multicolumn{1}{l|}{SG-Transformer + CogTree} & \textbf{22.9 / 28.4 / 31.0} &\multicolumn{1}{c|}{ 34.1 / 38.4 / 39.7 }  & \textbf{13.0 / 15.7 / 16.7} & \multicolumn{1}{c|}{20.8 / 22.9 / 23.4}  & \textbf{7.9 / 11.1 / 12.7} &15.1 / 19.5 / 21.7   \\ \hline
\end{tabular}}    
\caption{State-of-the-art comparison on R@K and mR@K. Our re-implemented SGG models are denoted by the superscript $*$.}
\label{tab:sota}
\end{table*}

\section{Experiments}
\label{sec:experiments}

\begin{figure}[t]
    \centering
    \includegraphics[width=\columnwidth]{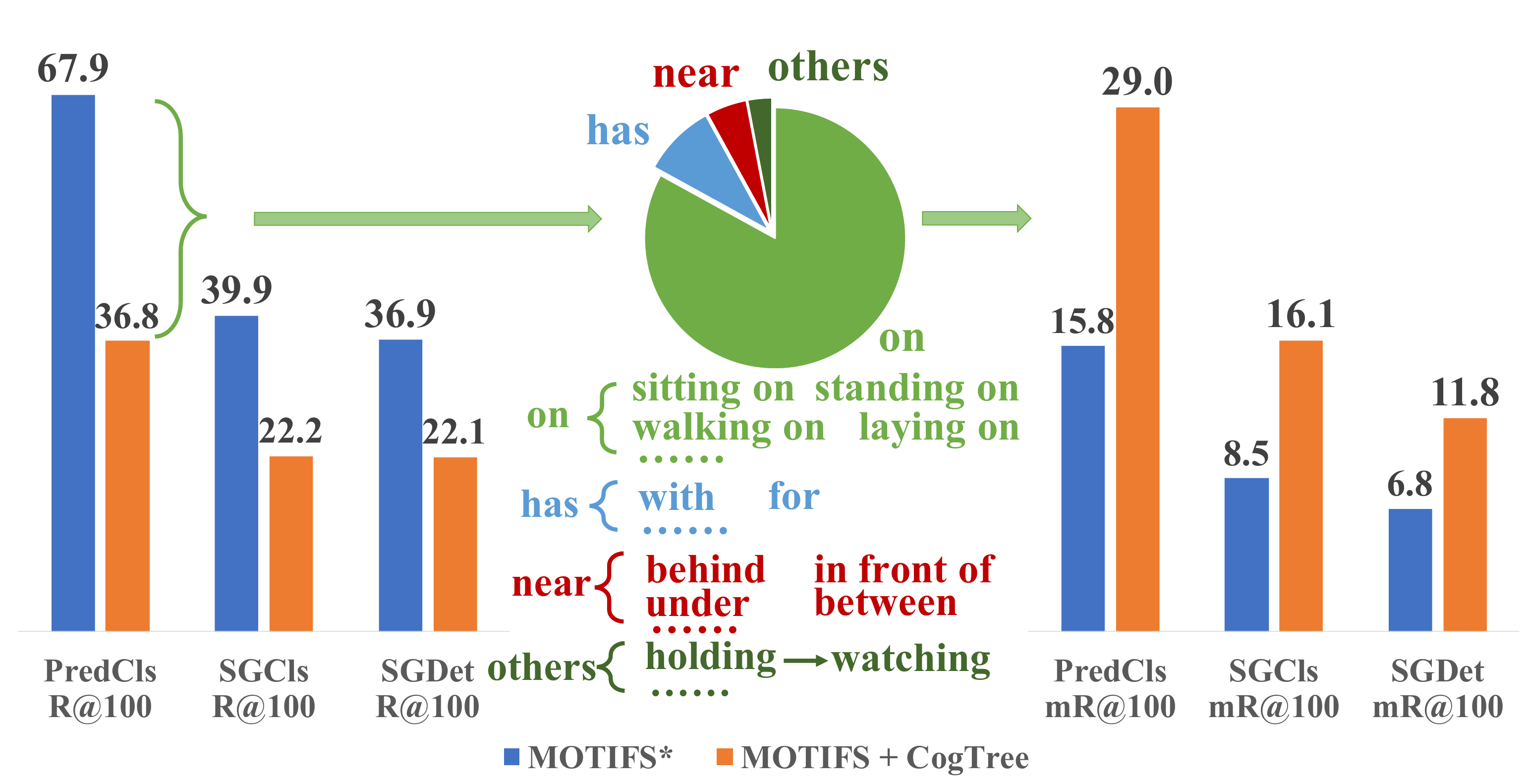}
    \caption{The pie chart shows the relationships that are correctly predicted by the baseline but regarded as incorrect by CogTree. The predictions by CogTree are shown under the pie chart. The increase of mR@K bringing a drop of R@K is commonly existed (see Table \ref{tab:sota}). Combining with the qualitative analysis, we believe that the contradiction of R@K and mR@K is caused by two reasons: (1) annotators prefer simple and vague  labels due to bounded rationality\protect\cite{simon1990bounded}; (2) CogTree predicts more fine-grained  relationships. Our analysis is consistent with the recent study \protect\cite{Tang2020Unbiased}.} 
    \label{fig:tsne}
\end{figure}

\textbf{Dataset:} 
We evaluate our models on 
the widely compared Visual Genome (VG) split \cite{xu2017scene},
with the 150 most frequent object classes and 50 most frequent relationship classes in VG \cite{krishna2017visual}. 
The VG split only contains training set and test set and we follow \cite{zellers2018neural} to sample a 5K validation set from the training set.
\\
\textbf{Tasks and Evaluation:} 
The SGG task contains three sub-tasks \cite{zellers2018neural}: (1) Predicate Classification (\textbf{PredCls}) takes the ground-truth object labels and bounding boxes for relationship prediction; (2) Scene Graph Classification (\textbf{SGCls}) takes ground-truth bounding boxes for object and relationship prediction; (3) Scene Graph Detection (\textbf{SGDet}) predict SGs from scratch. 
The conventional metric \textbf{Recall@K (R@K)} is reported but not considered as the main evaluation due to its bias. 
We adopt the recently proposed balanced metric \textbf{mean Recall@K  (mR@K)} \cite{chen2019knowledge,tang2019learning}. 
mR@K calculates R@K for each class independently, and then average the results.  
\\
\textbf{Implementation:} The object detector is the pre-trained Faster R-CNN \cite{ren2015faster} with ResNeXt-101-FPN \cite{Lin2017Feature}. 
$\lambda$ 
is set to 1 
and $\beta$ is set to 0.999. SG-Transformer contains 3 O2O blocks and 2 R2O blocks with 12 attention heads. Models are trained by SGD optimizer with 5 epochs. The mini-batch size is 12 and the learning rate is ${\rm 1.2\times10^{-3}}$.  Experiments are implemented with PyTorch and conducted with NVIDIA Tesla V100 GPUs.   

\subsection{State-of-the-Art Comparison}

We evaluate the CogTree loss 
on three models: MOTIFS, VCTree and SG-Transformer, and compare the performance with the state-of-the-art debiasing approach TDE \cite{Tang2020Unbiased}. 
We also compare with the biased models, including IMP+ \cite{xu2017scene}, FREQ \cite{zellers2018neural}, KERN \cite{chen2019knowledge}. 
The results are shown in Table \ref{tab:sota}. 
In the view of mR@K, CogTree remarkably improves all the baselines on all the metrics. The strongest baseline VCTree also achieves 3.6\%$\sim$13.6\% boost. 
CogTree consistently outperforms conventional debiasing methods: focal loss, reweight and resample, and the SOTA debiasing method TDE. 
Wherein, mR@20 achieves superior improvement compared with mR@50/100, which indicates that more accurate predictions have higher rank by CogTree. 
It is worth highlighting that CogTree outperforms TDE on both mR@K and most metrics of R@K on the strongest baseline VCTree.  \\
\textbf{Analysis of Labeling Bias. } 
We observe a drop of R@K  with the increase of mR@K from baselines to CogTree. Further analysis in Figure \ref{fig:tsne} reveals that annotators prefer simple and vague labels, which form the head classes. The baselines are biased on these head classes and result in high R@K.
CogTree prefers fine-grained and semantic-rich tail labels, causing the low R@K by ``incorrectly'' classifying the head classes into reasonable tail ones. 
To fairly compare with baselines on the conventional metric R@K, we compute R@K of the tail 45 classes from the predictions of all the classes. We observe a significant improvement on R@K from baselines to CogTree in Table \ref{tab:tailR@K}, which proves the strong debiasing ability of CogTree without sacrificing the performance of R@K.

\begin{table}[t]
\centering
\resizebox{\columnwidth}{!}{
\begin{tabular}{lccccccccc}
\hline
                               & \multicolumn{1}{c}{MOTIFS}                      & \multicolumn{1}{c}{VCTree}                           & \multicolumn{1}{c}{SG-transformer}       \\ \hline
 \multicolumn{1}{l|}{ Method }  & \multicolumn{1}{c|}{R@20 / 50 / 100}                          & \multicolumn{1}{c|}{R@20 / 50 / 100}                                    & R@20 / 50 / 100                                   \\ \hline
 \multicolumn{1}{l|}{baseline}    & \multicolumn{1}{c|}{35.0 / 36.7 / 37.0}                       & \multicolumn{1}{c|}{35.0 / 36.5 / 36.8}                               & 38.0 / 39.6 / 39.9                              \\ \hline
 \multicolumn{1}{l|}{baseline+CogTree}    & \multicolumn{1}{c|}{51.6 / 57.7 / 59.8}     & \multicolumn{1}{c|}{51.6 / 57.2 / 58.9}                               & 48.7 / 54.5 / 56.4                              \\ \hline
\end{tabular}}
\caption{R@K of the tail 45  classes  on predicate classification.}
\label{tab:tailR@K}
\end{table}

\begin{table}[t]
\centering
\resizebox{\columnwidth}{!}{\scriptsize
\begin{tabular}{llccccccccc}
\hline
&                                                                            & \multicolumn{1}{c}{PredCls}         & \multicolumn{1}{c}{SGCls}       & \multicolumn{1}{c}{SGDet}                                  \\ \hline
\multicolumn{2}{l|}{Method}                                                  & \multicolumn{1}{c|}{mR@20 / 50 / 100}                & \multicolumn{1}{c|}{mR@20 / 50 / 100}                & mR@20 / 50 / 100                                      \\ \hline
\multicolumn{2}{l|}{\textbf{Full loss}}           & \multicolumn{1}{c|}{\textbf{22.89 / 28.38 / 30.97}}  & \multicolumn{1}{c|}{\textbf{12.96 / 15.68 / 16.72}}  & \textbf{7.92 / 11.05 / 12.70}                        \\ \hline
\multicolumn{1}{l|}{1} & \multicolumn{1}{l|}{$\mathcal{L}_{TCB}$}  & \multicolumn{1}{c|}{21.08 / 27.08 / 29.41}           & \multicolumn{1}{c|}{12.15 / 15.07 / 16.15}           & 7.70 / 10.39 / 12.07                                 \\ 
\multicolumn{1}{l|}{2} & \multicolumn{1}{l|}{$\mathcal{L}_{CB}$}             & \multicolumn{1}{c|}{18.02 / 23.40 / 25.25}           & \multicolumn{1}{c|}{10.76 / 13.13 / 13.88}           & 6.74 / 9.56 / 11.29                                  \\ \hline
\multicolumn{1}{l|}{3} & \multicolumn{1}{l|}{$\mathcal{L}_{TCE}$}  & \multicolumn{1}{c|}{21.16 / 26.14 / 28.32}           & \multicolumn{1}{c|}{12.14 / 14.42 / 15.29}           & 7.57 / 10.53 / 11.86                                 \\
\multicolumn{1}{l|}{4} & \multicolumn{1}{l|}{ $\mathcal{L}_{CE}$}            & \multicolumn{1}{c|}{14.35 / 18.48 / 20.21}           & \multicolumn{1}{c|}{8.57 / 11.46 / 12.27}            & 5.55 / 7.74 / 8.98                                   \\  \hline
\multicolumn{1}{l|}{5} & \multicolumn{1}{l|}{Fuse-subtree}   & \multicolumn{1}{c|}{16.20 / 20.17 / 22.12}           & \multicolumn{1}{c|}{8.71 / 10.66 / 11.61}            & 5.36 / 7.19 / 8.28                                   \\
\multicolumn{1}{l|}{6} & \multicolumn{1}{l|}{Fuse-layer}     & \multicolumn{1}{c|}{13.77 / 18.87 / 20.77}           & \multicolumn{1}{c|}{8.17 / 10.39 / 11.32}            & 5.86 / 8.02 / 9.05                                   \\ \hline
\multicolumn{1}{l|}{7} & \multicolumn{1}{l|}{$\mathcal{L}$(MAX)}   & \multicolumn{1}{c|}{15.48 / 19.93 / 21.87}           & \multicolumn{1}{c|}{8.97 / 10.85 / 11.83}            & 5.38 / 7.16 / 8.16                                   \\ 
\multicolumn{1}{l|}{8} & \multicolumn{1}{l|}{$\mathcal{L}$(SUM)}   & \multicolumn{1}{c|}{11.31 / 15.67 / 17.98}           & \multicolumn{1}{c|}{6.58 / 8.82 / 9.86}              & 1.86 / 3.09 / 3.68                                   \\  \hline
\end{tabular}}
\caption{Ablation study of key components in CogTree loss.}
\label{tab:ablation}
\end{table}

\begin{figure*}[t]
    \centering
    \includegraphics[width=\textwidth,height=0.41\textwidth]{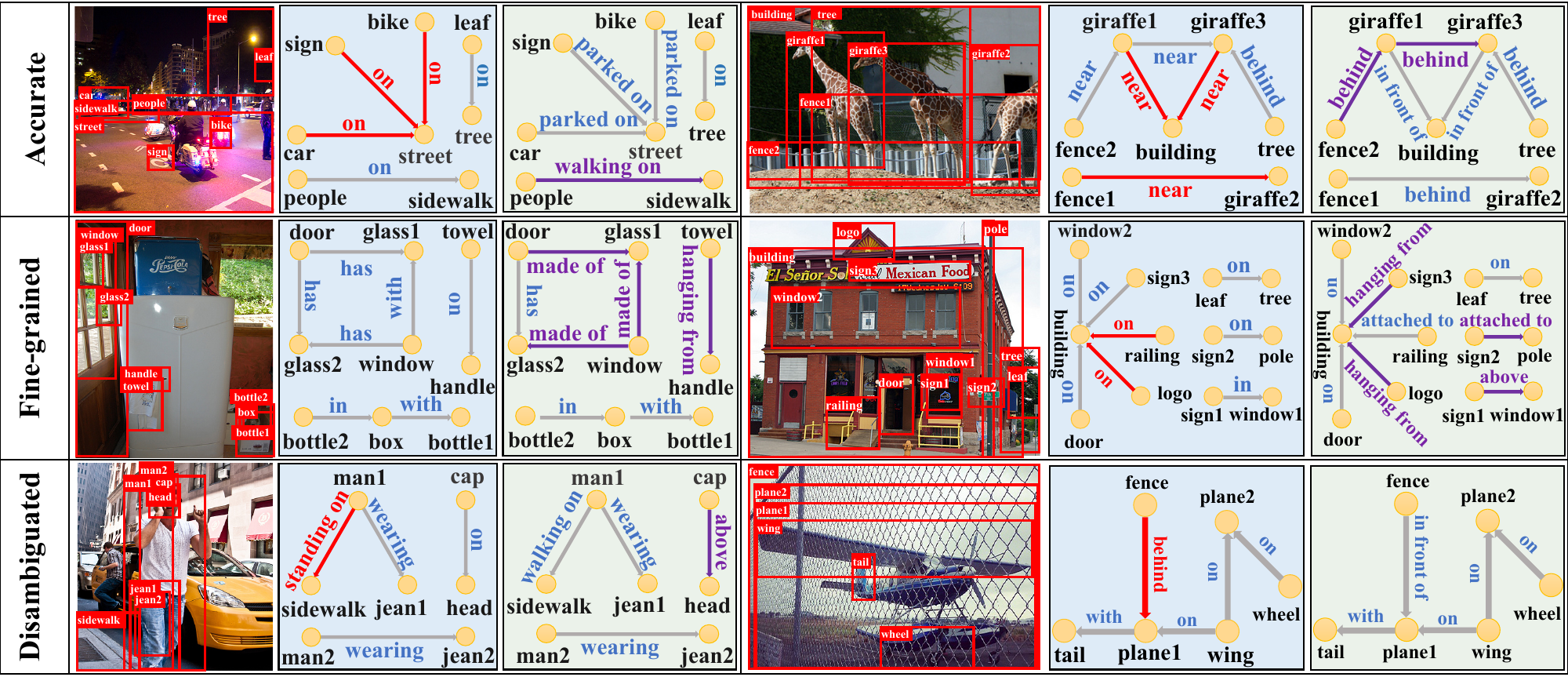}
    \caption{Visualization of scene graphs generated by SG-Transformer (blue) and SG-Transformer+CogTree (green). Compared with the ground-truth, the quality of predicted relationships are marked in three colors: red (false), blue (correct), purple (better).}
    \label{fig:cases}
\end{figure*}

\subsection{Ablation Study}
We investigate each loss terms, the weighting method,
two tree building strategies  and two methods for computing PCP and CBW: 1) $\mathbf{\mathcal{L}_{TCB}}$:
tree-based classficition loss term in Eq. \ref{eq:TCB}; 2) $\mathbf{\mathcal{L}_{CB}}$: flat classification loss term in Eq. \ref{eq:CB}; 3) $\mathbf{\mathcal{L}_{TCE}}$: we remove the weights $\{w_i\}$ from Eq. \ref{eq:TCB}; 4)  $\mathbf{\mathcal{L}_{CE}}$: we remove the weight $w_k$ from Eq. \ref{eq:CB}; 5) \textbf{Fuse-subtree}: we mix  relationships  belonging  to  different  concepts in one subtree; 6) \textbf{Fuse-layer}: we 
mix the fine-grained  and coarse relationships  in  the  same layer of
each subtree; 7) $\mathbf{\mathcal{L}}$\textbf{(MAX)}: we use MAX instead of AVERAGE to compute $z_i$ and $w_i$ for non-leaf nodes in Eq.\ref{eq:probability} and Eq.\ref{eq:weight}; 8) $\mathbf{\mathcal{L}}$\textbf{(SUM)}: we use SUM instead of AVERAGE. 

The results in Table \ref{tab:ablation} on the SG-Transformer baseline indicate that: 1) $\mathcal{L}_{TCB}$ and $\mathcal{L}_{CB}$ have complementary benefits for SGG, where $\mathcal{L}_{TCB}$ has obvious greater influence. 2) Both of the loss terms benefit from the weighting strategy. 
3) The CogTree is built based on two principles: a) Relationships belonging to the same concept are organized in one subtree and b) relationships in one subtree are organized in different layers from coarse to fine. A significant performance decrease 
is observed when the tree violates either of the principles. 4) AVERAGE performs the best because that both MAX and SUM increase the predicted probabilities and decrease the incorrect prediction penalty in Eq. \ref{eq:TCB}.

\subsection{Qualitative Analysis}
We visualize several PredCls samples in Figure \ref{fig:cases}. SG-Transformer achieves obvious improvement when equipped with CogTree: (1) CogTree predicts more fine-grained relationships. As shown in the first two rows, the baseline mostly predicts the vague head classes  {\ttfamily on} and  {\ttfamily near} 
while CogTree predicts actions and locations like  {\ttfamily parked on} 
and {\ttfamily in front of}. 
(2) CogTree accurately distinguishes visually and semantically similar relationships as shown in the last row. However, the baseline falsely predicts  {\ttfamily walking on} as {\ttfamily standing on} and predicts {\ttfamily in front of} as {\ttfamily behind}, since  flat classification cannot capture the detailed discrepancy. 
In summary, CogTree preferences towards fine-grained relationships instead of the vaguely biased ones.



\begin{table}[t]
\centering
\resizebox{\columnwidth}{!}{\scriptsize
\begin{tabular}{cccccccccc}
\hline
                                & \multicolumn{1}{c}{Predicate Classification}         & \multicolumn{1}{c}{Scene Graph Classification}           & \multicolumn{1}{c}{Scene Graph Detection}                              \\ \hline
\multicolumn{1}{c|}{$\lambda$}  & \multicolumn{1}{c|}{mR@20 / 50 / 100}                & \multicolumn{1}{c|}{mR@20 / 50 / 100}                    & mR@20 / 50 / 100                                  \\ \hline
\multicolumn{1}{c|}{0.4}        & \multicolumn{1}{c|}{21.17 / 27.20 / 29.59}           & \multicolumn{1}{c|}{12.09 / 15.03 / 16.12}               & 7.22 / 9.72 / 11.30                                \\
\multicolumn{1}{c|}{0.7}        & \multicolumn{1}{c|}{21.28 / 26.39 / 28.69}           & \multicolumn{1}{c|}{12.22 / 14.91 / 15.88}               & \textbf{8.62 / 11.30 / 12.70}                      \\
\multicolumn{1}{c|}{1}          & \multicolumn{1}{c|}{\textbf{22.89 / 28.38 / 30.97}}  & \multicolumn{1}{c|}{\textbf{12.96 / 15.68 / 16.72}}      & 7.92 / 11.05 / 12.70                              \\
\multicolumn{1}{c|}{1.3}        & \multicolumn{1}{c|}{22.06 / 27.11 / 29.21}           & \multicolumn{1}{c|}{12.35 / 15.04 / 15.87}               & 7.84 / 10.48 / 12.19                               \\
\multicolumn{1}{c|}{1.6}        & \multicolumn{1}{c|}{21.34 / 26.80 / 29.18}           & \multicolumn{1}{c|}{12.20 / 14.70 / 15.82}               & 6.78 / 9.04 / 10.22                                \\\hline
\end{tabular}}
\caption{Performance with different balancing weights.}
\label{tab:parameter-loss}
\end{table}

\subsection{Parameter Analysis}
In Table \ref{tab:parameter-loss}, we assess the effect of $\lambda$ in Eq.\ref{eq:fullLoss}. $\lambda=1$ achieves the best performance on most metrics. The performance drops slightly in the range of $[0.7,1.3]$. We also vary the number of O2O and R2O blocks from 2 to 4 in SG-Transformer and obtain the highest mR@K with 3 O2O blocks and 2 R2O blocks. We use the above settings in our full model.


\section{Conclusion}
In this paper, we propose a CogTree loss to generate  unbiased scene graphs with highly biased data. 
We novelly leverage the biased prediction from SGG models to organize the independent relationships by a tree structure, which contains multiple layers corresponding to the relationships from coarse to fine. We propose a CogTree loss specially for the above tree structure that supports hierarchical distinction for the correct relationships while progressively eliminating the irrelevant ones. The loss is model-agnostic and consistently boosting the performance of various SGG models with remarkable improvement. How to incorporate commonsense knowledge to optimize the CogTree structure will be our future work.

\section*{Acknowledgements}
This  work  was  supported  by  the  National  Natural  Science  Foundation  of China (Grant No. 62006222).

\bibliography{ijcai2021}
\bibliographystyle{named}
\end{document}